\documentclass[]{interact}

\usepackage{epstopdf}
\usepackage[caption=false]{subfig}

\usepackage[numbers,sort&compress]{natbib}
\bibpunct[, ]{[}{]}{,}{n}{,}{,}
\renewcommand\bibfont{\fontsize{10}{12}\selectfont}
\makeatletter
\def\NAT@def@citea{\def\@citea{\NAT@separator}}
\makeatother

\usepackage[utf8]{inputenc}
\usepackage{url}
\usepackage{graphicx}
\graphicspath{{images/}}
\usepackage{csquotes}
\usepackage{booktabs}
\usepackage{array}
\usepackage{amssymb}
\usepackage{tabularray}
\usepackage{tabularx}
\usepackage{makecell}
\usepackage[title]{appendix}

\usepackage{pdflscape}
\usepackage{afterpage}

\theoremstyle{plain}

\theoremstyle{definition}

\theoremstyle{remark}

\newcommand{\tc}{T$_{c}$}
\providecommand{\keywords}[1]

\begin{document}

\articletype{RESEARCH PAPER}

\title{Automatic extraction of materials and properties from superconductors scientific literature}

\author{
    \name{Luca Foppiano\textsuperscript{a}\thanks{Corresponding authors: Luca Foppiano (luca@foppiano.org) and Masashi Ishii (ISHII.Masashi@nims.go.jp)}, Pedro Baptista de Castro\textsuperscript{b}, Pedro Ortiz Suarez\textsuperscript{c}, Kensei Terashima\textsuperscript{b}, Yoshihiko Takano\textsuperscript{b}, Masashi Ishii\textsuperscript{a}}
    \affil{\textsuperscript{a}Material Database Group, MaDIS, NIMS, Tsukuba, JP; \textsuperscript{b}Nano Frontier Superconducting Materials Group, MANA, NIMS, Tsukuba, JP; \textsuperscript{c}Data and Web Science Group, University of Mannheim, Mannheim, DE}
}

\maketitle

\begin{abstract}
    The automatic extraction of materials and related properties from the scientific literature is gaining attention in data-driven materials science (Materials Informatics).
    In this paper, we discuss Grobid-superconductors, our solution for automatically extracting superconductor material names and respective properties from text.
    Built as a Grobid module, it combines machine learning and heuristic approaches in a multi-step architecture that supports input data as raw text or PDF documents.
    Using Grobid-superconductors, we built SuperCon\textsuperscript{2}, a database of 40324 materials and properties records from 37700 papers.
    The material (or sample) information is represented by name, chemical formula, and material class, and is characterized by shape, doping, substitution variables for components, and substrate as adjoined information.
    The properties include the T\textsubscript{c} superconducting critical temperature and, when available, applied pressure with the T\textsubscript{c} measurement method.
\end{abstract}



\section{Introduction}
In recent years, with the creation of computational databases, such as the Materials Project (MP)~\cite{materialsprojectJain2013} and the Open Quantum Materials Database (OQMD)~\cite{oqmdkirklin2015open}, and then experimental data repositories such as NIMS MDR (\url{http://mdr.nims.go.jp})~\cite{ranganathan_anusha_2019_3553963}, focus has been steadily shifting towards a data-driven design of materials, which is often called Materials Informatics (MI).
Such an approach is expected to accelerate the exploration of functional materials because it is not limited to the intuition or experience of very little genius researchers.
In this new paradigm, the efficient use of data to guide experiments and material property prediction through the use of machine learning methods takes center stage.
For example, data-driven methods have been used to search/design magneto-caloric materials~\cite{Bocarsly2017,Castro2020-12,court2021inverse}, photo-catalysts for hydrogen splitting~\cite{xiong2021optimizing}, thermoelectrics~\cite{iwasaki2019machine}, and superconductors~\cite{stanev_machine_2017}.
In such a data-driven search, one of the most important keys lies in the availability of the data, which should at least should consist of compositions of materials and their physical properties.
In the specific case of superconductivity, most of the data-driven works~\cite{stanev_machine_2017, le2020critical,Hamlin2019SuperconductivityNR} rely on a single database: SuperCon (\url{http://supercon.nims.go.jp}).

SuperCon is a structured database of superconductor materials and properties;it was developed at the National Institute for Materials Science (NIMS) in Japan.
At the time of writing this paper, SuperCon contained about 33000 inorganic and 600 organic materials and is the ``de-facto'' standard in data-driven research for superconductors materials (about 4400 articles contain the mention ``\textit{SuperCon database}'' in Google Scholar).
However, the SuperCon harvesting process is currently fully manual ``from scratch'': humans have to read the human-readable printed matter such as PDF documents and enter the information into the system.
The efficiency is directly proportional to the number of available human curators.
Considering the cost of database construction, it is necessary to consider an assisted or alternative system that improves throughput while ensuring data quality equivalent to that of manual extraction.

As a solution, we are developing a hybrid data extraction method from scientific literature that combines automation using text data mining and manual curation.
The automated system extracts and formats potential data and proposes them to the curator as ``pre-cooked'' structured data by (1) highlighting the relevant entities on the original document and (2) pre-filling the extracted information in a tabular format.
In building the automated part of this hybrid system (training, evaluation) we used SuperMat, an annotated linked dataset from scholarly documents in superconductor research, which we recently constructed~\cite{foppiano2021supermat}.

In this work, we present Grobid-superconductors: a system that automatically extracts structured information related to superconductor materials and properties from scientific literature.
The tool is a specialised module of Grobid (Generation of Bibliographic Data)~\cite{GROBID}, a machine learning library designed to parse and structure scientific documents.
Grobid provides an open-source platform for building specialised modules including astronomical entities recognition\cite{grobid-astro}, dictionaries~\cite{khemakhem:hal-01508868}, software mentions~\cite{lopez2021mining}, and physical measurements extraction~\cite{foppiano2019quantities}.
Grobid provides several built-in features including access to PDF document layout information, citation resolution, bibliographic information consolidation through biblio-glutton~\cite{biblio-glutton-lookup} (a fast open-source reference matching service for CrossRef data), and a diverse set of machine learning (ML) architectures from a fast linear Conditional Random Field (CRF) to the latest state-of-the-art deep learning implementations.

Using Grobid-superconductors and other sub-tools, we established a pipeline to process a large number of documents and obtain an automated database of superconductor materials and properties.
We processed 37770 papers from ArXiv (\url{https://arxiv.org}) and obtained a database of 40324 records.
This new database, named SuperCon\textsuperscript{2}, can become an automated staging area for SuperCon, when bridged by a curation interface.
During the project, there was also an opportunity to focus on properties that that have recently become of greater interest and that are underrepresented in SuperCon.
For example, the ``pressure'' applied to obtain superconductivity (about 20 records in SuperCon), has gained attention because it can radically change the physical structure of a material.
In addition, the ``method'' used to measure the superconducting transition temperature \tc~(about 600 records in SuperCon) can be used to semantically recognise multiple \tc 's obtained from the same material or sample (e.g., distinguish calculated and experimental values of \tc).

\section{Grobid-superconductors}

\begin{figure}[ht]
    \includegraphics[width=\textwidth]{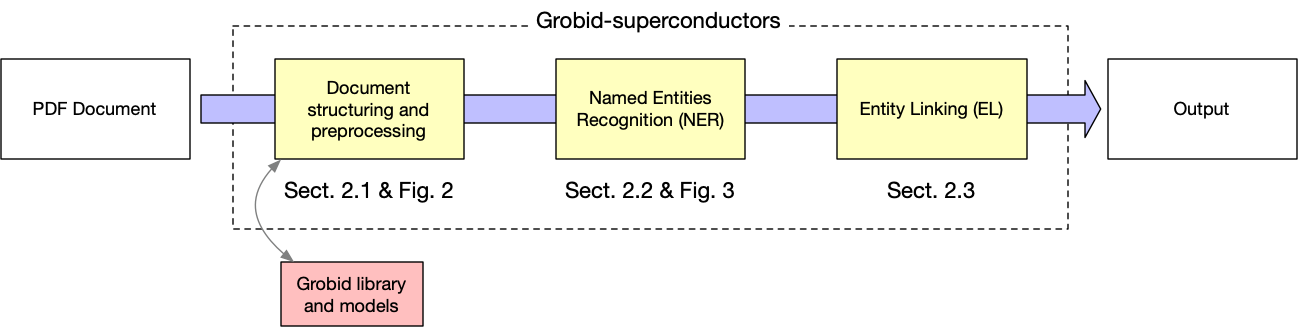}
    \caption{Processing pipeline for extracting superconductors materials and properties. }
    \label{fig:pipeline-overview}
\end{figure}

We developed Grobid-superconductors as a Grobid module following principles (multi-step, sentence-based, full-text-based) discussed in a previous preliminary study~\cite{foppiano:hal-02870896}.
Grobid has several advantages: 1) it can be integrated with pdfalto (\url{https://github.com/kermitt2/pdfalto}), a specialised tool for converting PDF to XML, which mitigates extraction issues such as the resolution of embedded fonts, invalid character encoding, and the reconstruction of the correct reading order, 2) it allows access to PDF document layout information for both machine learning and document decoration (e.g., coordinates in the PDF document); and, 3) it provides access to a set of high-quality, pre-trained machine learning models for structuring documents.
Grobid-superconductors is structured as a three-steps process illustrated in Figure~\ref{fig:pipeline-overview} and described in the Sections~\ref{subsubsec:document-structuring},~\ref{subsubsec:extraction}, and~\ref{subsubsec:linking}.

\paragraph*{Abstract versus full-text}
At the time of writing this paper, we are aware of related works that utilise text from abstracts as training data for machine learning.
The main reason for using abstracts is that they are usually freely available as text~\cite{kononova_text-mined_2019}, and contain condensed information~\cite{yamaguchi-etal-2020-sc, court_magnetic_2020}.
Accurately parsing the full-text presents more challenges, however, but they are mitigated by Grobid and, the full-text contain a broader range of information, including the sample preparation process, negative results (e.g., absence of superconductivity for certain samples), and background information (e.g., reports on other materials from referenced works).
Thus, grobid-superconductors is built to support full-text documents.

\paragraph*{Paragraphs versus sentences}
Another question related to natural language processing (NLP) is whether to use sentence-based or paragraph-based text.
While paragraphs can be extracted as part of the layout of PDF documents, obtaining sentences adds an additional step in which text is processed with a sentence segmenter.
However, sentences are almost always shorter by definition, and in deep learning, this has advantages.
In training and prediction, sentences will likely be shorter than the ``max sequence length'' limitation (e.g., 512 tokens for transformers).
During training, sentences also use less memory and allow us to train models with a larger ``batch size'', which has been shown to improve efficiency and obtain better results~\cite{roberta}. 

We chose to use sentence-based text in Grobid-superconductors after performing preliminary experiments on our tasks typologies, but on a smaller scale. 
For the Named Entities Recognition (NER) task we trained and evaluated a sequence labelling model for each version (paragraph-based and sentence-based) on four annotated documents (3/1 document partition for training/evaluation) from SuperMat~\cite{foppiano2021supermat}.
As indicated in Table~\ref{tab:comparison-evaluation-sentences-paragraphs}, the F1-score increased by 17.94 percentage points when the sentence-based text was used.

For the Entity Linking (EL) task, we want to maximise precision. 
In our previous work~\cite{foppiano2019proposal} we noticed that limiting linking entities within the same sentence (versus paragraph) would obtain higher precision (68.7\% versus 57\%) at the expense of lower recall (6.5\% versus 10.7\%), and F1-score (11.87\% versus 18.01\%).
Therefore, in both our tasks we found evidence that a sentence-based dataset is more beneficial than paragraph-based dataset.

\subsection{Document structuring and pre-processing}
\label{subsubsec:document-structuring}
\begin{figure}[ht]
    \includegraphics[width=\textwidth]{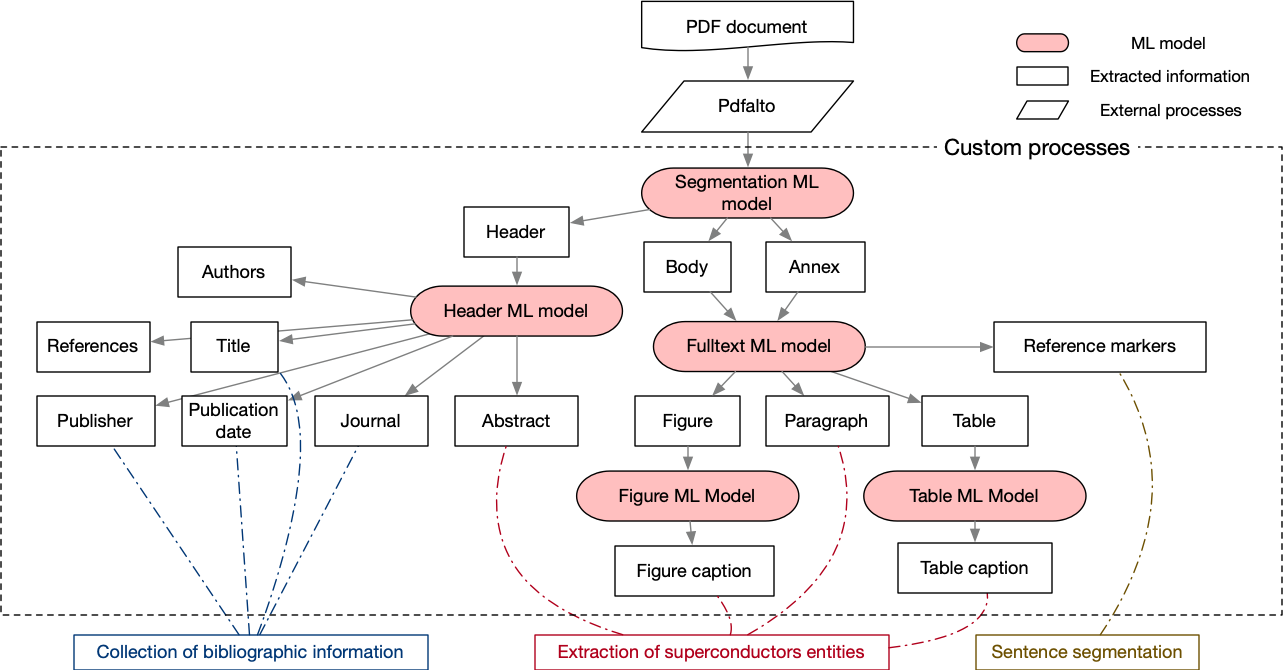}
    \caption{Grobid-superconductors extraction processes (bibliographic information, superconductor entity extraction and sentence segmentation) within the Grobid cascade data flow.}
    \label{fig:grobid-document-processing}
\end{figure}

In the first step of our process, the PDF document is converted into an internal model based on a list of text statements, tokens, and features.
The input document is processed using the Grobid original models, where we apply customised processes for document header and content.
We select a subset of bibliographic information from the header: title, authors, DOI, publisher, journal, and year of publication, and we consolidate them via Grobid to match the publisher's quality (even by processing the ``preprint version'' of the publication).
The superconductors entities extraction is applied to the content, only on relevant text items: title, abstract, text content from body or annexes, text content from figure and table captions (Figure~\ref{fig:grobid-document-processing}).

We use the collected reference markers (also called \textit{reference callouts}) from the text as features for improving the paragraph segmentation in sentences: the segmentation is cancelled if the end of sentence falls within the boundaries of a reference marker.
For example, a sentence containing a reference in the form ``Foppiano et. al.'' may be mistakenly segmented in the middle at the token ``et.''.


\subsection{Named Entity Recognition}
\label{subsubsec:extraction}

In the second step, the Named Entity Recognition (NER) task is performed on the previously extracted text.

\subsubsection{Overview}

\begin{figure}[ht]
    \includegraphics[width=\textwidth]{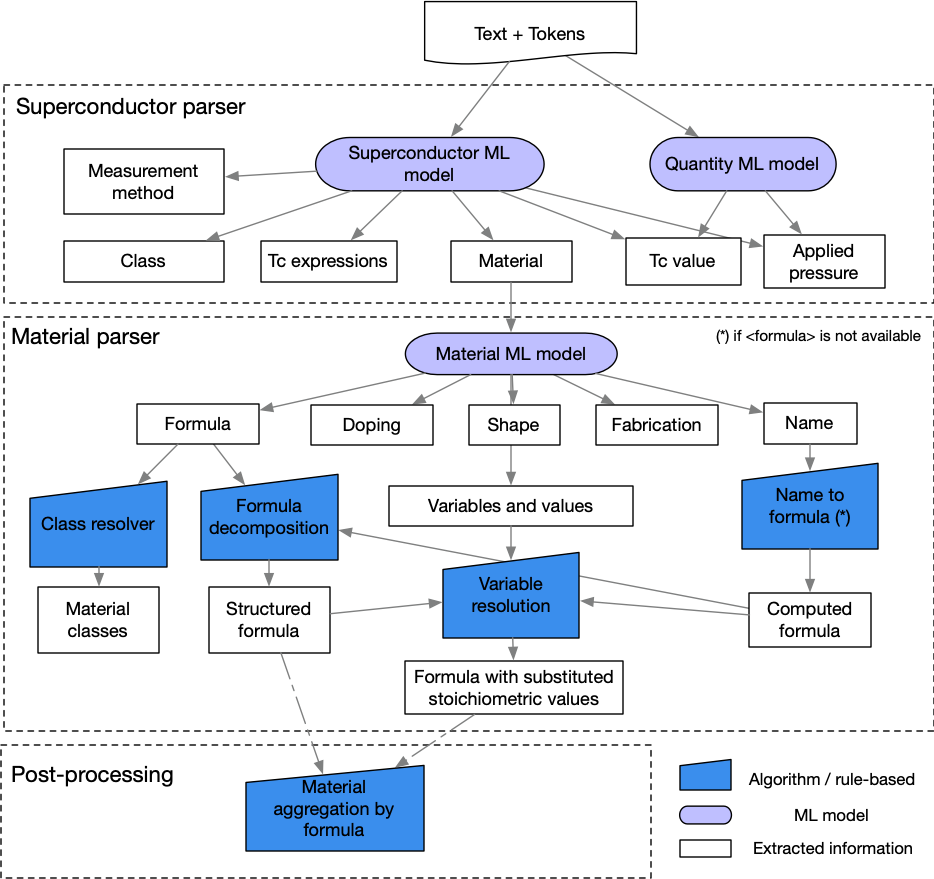}
    \caption{\label{fig:extraction-ml-models-cascade-architecture} Cascade architecture in the NER step. The white rectangles indicate the extracted information (described in Tables~\ref{tab:superconductors-parser-entities} and~\ref{tab:material-parser-entities}).}
\end{figure}

As illustrated in Figure~\ref{fig:extraction-ml-models-cascade-architecture} the ``superconductor parser'' extracts the main superconductor-related information by aggregating the resulting entities from two ML models.
The Superconductors ML model was developed based on the SuperMat schema~\cite{foppiano2021supermat}, and the Quantity ML model was developed in a separated Grobid module for measurement extraction~\cite{foppiano2019quantities} and the output is limited to only temperatures and pressures.
Overlapping entities are merged, exacted duplicates are removed, and the largest entities (in terms of string length) are preserved.
The resulting entities are summarised in Table~\ref{tab:superconductors-parser-entities}.

Entities of type \texttt{<material>}, which may contain mixed heterogeneous information, are passed in the cascade to the ``Material parser'' which aggregates ML and other tools.
First, the entity is passed through a Material ML model to segment and identify its content (Table~\ref{tab:material-parser-entities}).
Then, different processes are applied, depending on which information is available. 
These processes include the following:
\begin{itemize}
    \item Formulas are decomposed into a structured composition. We identify each element-stoichiometry pair (e.g., ``O'': 7.0) using mat2chem~\cite{kononova_text-mined_2019} and Pymatgen~\cite{Ong2013}; if only the material name is available, we lookup its formula (e.g., hydrogen to \textit{H}),
    \item Using heuristics, we classify the formula by assigning multiple classes as they are understood from superconductor researchers, for example cuprate, oxides, alloys, etc.
    \item Using the variables and values extracted, we substitute them into partial formulas. For example, in \texttt{La 4 Fe 2 A 1-x O 7 (A=Mg,Co; x=0.1,0.2)}, we substitute \textit{A} and \textit{x} using their parsed values, and applying permutations, we obtain four \textit{resolved formulas}: \texttt{La 4 Fe 2 Mg 0.9 O 7}, \texttt{La 4 Fe 2 Mg 0.8 O 7}, \texttt{La 4 Fe 2 Co 0.9 O 7}, and \texttt{La 4 Fe 2 Co 0.8 O 7}.
\end{itemize}

Finally, after all entities are extracted, the post-processing aggregates different mentions of the same materials using the parsed formulas at the document-level.
For example, formula with partial substitutions such as \texttt{La 2 Fe 1-x O 7 (x = 0.1, 0.2)} will be aggregated with materials like \texttt{La 2 Fe 0.9 O 7} appearing in other sections of the same document.

\subsubsection{Machine Learning study}

In this section we discuss the novel ML models we have trained for extracting specialised entities: the Superconductor ML model and the Material ML model (Figure~\ref{fig:extraction-ml-models-cascade-architecture}).
SuperMat~\cite{foppiano2021supermat}, our training dataset, contains 164 papers as of the time of writing and is composed of annotated full-text and layout features from PDF documents.

For both ML models we trained and evaluated the following four architecture/implementations: linear CRF (CRF), bidirectional LSTM with CRF~\cite{Lample2016NeuralAF} (BidLSTM\_CRF), bidirectional LSTM with CRF with Features~\cite{Lample2016NeuralAF} (the same as (BidLSTM\_CRF) with an additional input channel for features; BidLSTM\_CRF\_FEATURES), and SciBERT~\cite{Beltagy2019SciBERT} using a CRF as the activation layer (Scibert).

The ML models are interfaced by Grobid, which uses the Wapiti\cite{lavergne2010practical} implementation for linear CRF, and DeLFT (Deep Learning For Text)~\cite{DeLFT} for deep learning models.
The architectures CRF and BidLSTM\_CRF\_FEATURES make use of the orthogonal features we have summarised in Table~\ref{tab:ML-model-features}.

\paragraph*{Superconductor ML model}

\begin{figure}[ht]
    \centering
    \includegraphics[width=\textwidth]{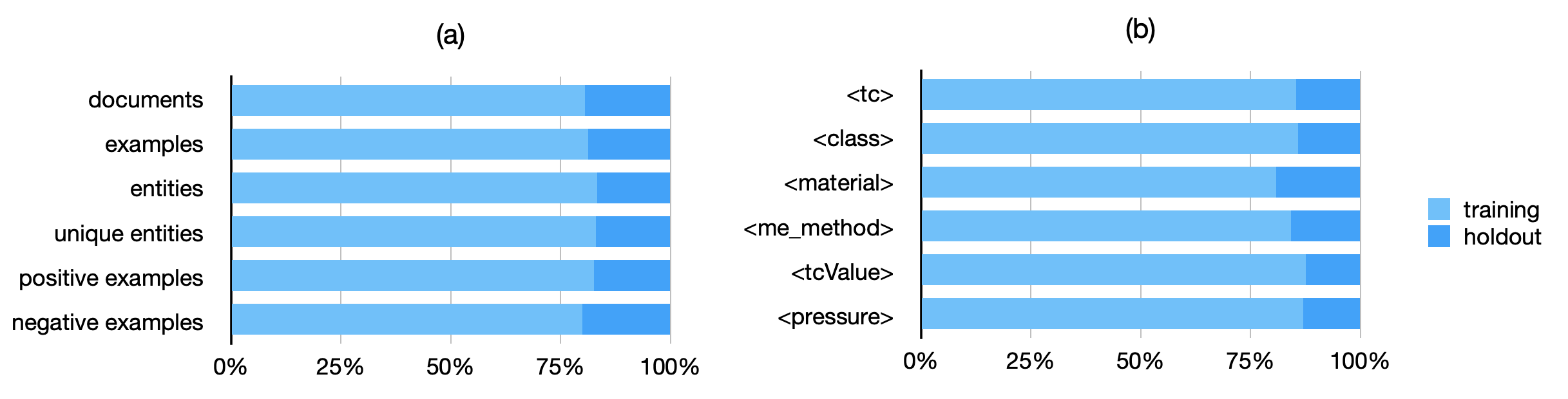}
    \caption{Holdout/training set distribution for (a) general metrics and (b) entity labels; entities and unique entities indicate the number of labelled entities with and without value duplicates, respectively, and positive examples (+) and negative examples (-) indicate the number of sentences with at least one entity and with no entities, respectively.}
    \label{fig:training-holdout-set-distribution}
\end{figure}

\subparagraph*{Holdout set}
The holdout set evaluation consists in using a fixed part of a dataset for validation. 
The selection must be performed to reproduce the same distribution of entities of the original dataset.
We assembled the holdout set by manually selecting 32 documents (24\%) from SuperMat, making sure they had a similar ratio of examples, entities and unique entities with the remaining 76\% (132 documents) which was used as training set (Figure~\ref{fig:training-holdout-set-distribution}a).
Maintaining the same rate for entity type distribution between the two sets was more challenging: on average, we obtained about 15-18\% of labels of each type in the holdout set (Figure~\ref{fig:training-holdout-set-distribution}b), except for the \texttt{<material>} label (23\%). 

\begin{figure}[ht]
    \centering
    \includegraphics[width=0.6\textwidth]{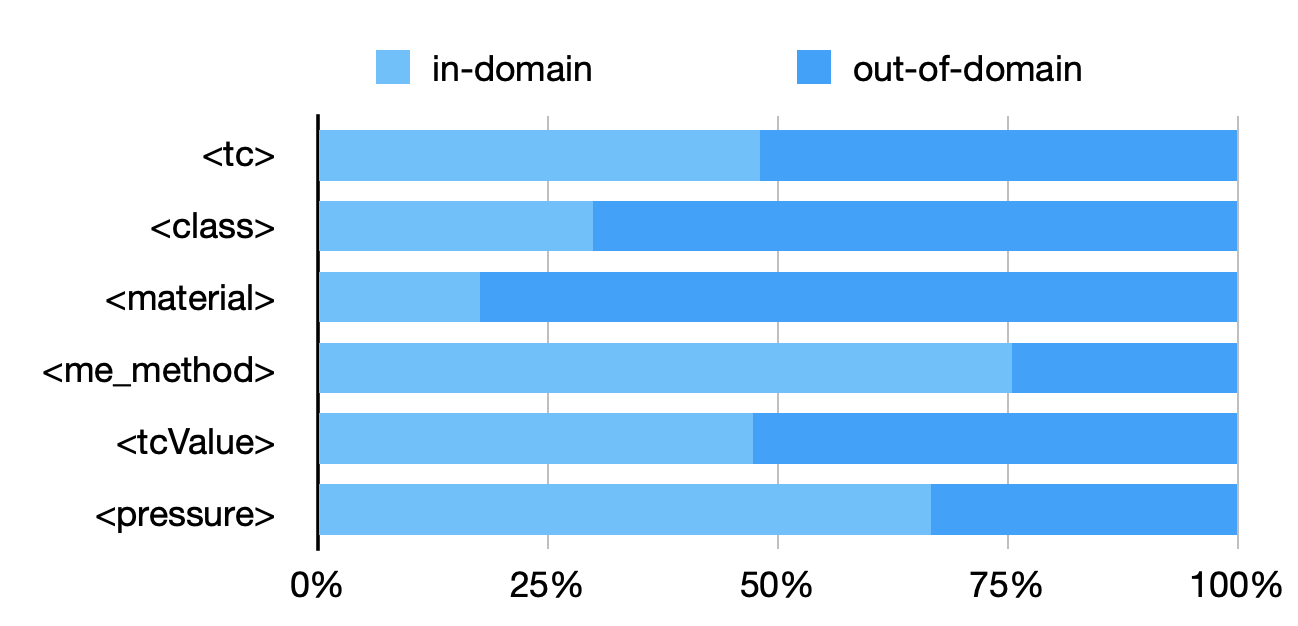}
    \caption{Holdout ``out-of-domain'' rates. The entities from the holdout set that are also in the training set are ``in-domain'' and the entities that are not in the training set are ``out-of-domain''.}
    \label{fig:out-domain-holdout}
\end{figure}

We defined the ``out-of-domain'' ratio as the number of unique entities from the holdout set that were not in the training set.
The holdout set ``out-of-domain'' ratio was on average around 72\%, which challenge the model generalisation (every 100 entities in the holdout set, 72 were never seen before during training).
Most of the labels had an ``out-of-domain'' ratio above 50\%  (Figure~\ref{fig:out-domain-holdout});  \texttt{<material>}, the most important label, had the highest ratio (82\%) while \texttt{<me\_method>} and \texttt{<pressure>} have the lowest (25\% and 33\%). 
The low ratio of \texttt{<me\_method>} can be explained by their low entity variability (11.44\%).

\subparagraph*{Positive sampling}
We trained the model with positive sampling by removing the examples without entities (negative examples, Figure~\ref{fig:training-holdout-set-distribution}a).
This approach provided an improvement of 2\% in both precision and recall as compared to the result without sampling when testing against the holdout set.
Additional experiments with active and random sampling~\cite{lopez2021mining} with ratios of negative examples of 0.1, 0.25, 0.5 and 1.0 did not provide stable evidence suggesting scoring improvements when testing against the holdout set. 

\subparagraph*{Evaluation}
The best results were obtained by Scibert with an F1 of 77.03\% and a recall of around 80.69\% (Table~\ref{tab:evaluation-superconductors-ML-model}).
The features did not provide any improvements with RNN models: BidLSTM\_CRF and BidLSTM\_CRF\_FEATURES resulted in the same F1 score.
This result comes as a surprise because features such as superscript/subscript were expected to be determinants for recognising material sequences.

The \texttt{<pressure>} label had the lowest performance scores in all architectures. We believe that 274 training examples are not a sufficient large number considering that pressure expressions can be dependent on the context because they can refer to different types of pressures (e.g., annealing pressure).
The label with the highest score was \texttt{<material>}, with F1 values of 80.77\% and 78.06\% for Scibert and BidLSTM\_CRF, respectively. In addition, \texttt{<material>} had the highest ``out-of-domain'' ratio in the holdout set (greater than 75\%, Figure~\ref{fig:out-domain-holdout}) and the highest ``label variability'' (the ratio between unique entities and total entities, about 42\%), which suggests that the model recognises correctly materials that has not been ``seen'' during the training.
On the other hand, the \texttt{<me\_method>} label, which has lower ``label variability'' (around 11\%) and a low ``out-of-domain'' ratio, had an F1 score of 66.56\% with Scibert and 65.92\% with BidLSTM\_CRF.
For \texttt{<tc>}, the CRF outperformed the other architectures (F1 score of 83.96\%), especially Scibert (78.35\%). 
This outcome can be explained by the extremely low variability (12.69\%) of entities labelled as \texttt{<tc>}. 

\begin{figure}[ht]
    \centering
    \includegraphics[width=\textwidth]{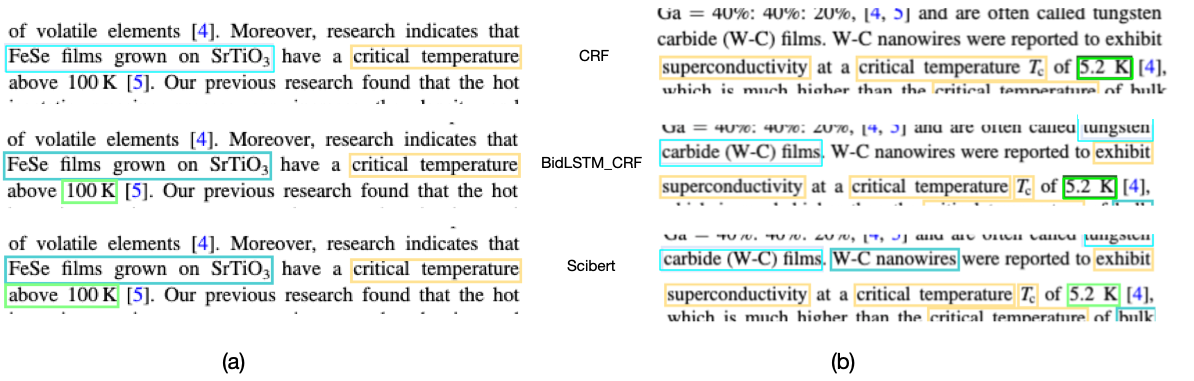}
    \caption{Examples taken from two sources~\cite{Gajda_2016, Shibata_2016} of results from three different architectures: CRF, BidLSTM\_CRF and, Scibert. The boxes annotating the text represent the extracted entities (material are indicated in light blue, \tc in green, and \tc expressions in yellow).}
    \label{fig:example-comparison-architectures}
\end{figure}

Scibert shows good generalisation capacity for unseen examples or examples appearing in different contexts.
For example, in Figure~\ref{fig:example-comparison-architectures}a, only Scibert correctly extracts ``above 100K'', while CRF misses it completely and BidLSTM\_CRF misses ``above''.
In the training data, ``above 100K'' is not present, but ``below 100K'' and ``\~100K'' are present, and several other entities contain the token ``above'' and Scibert can understand that the token ``above'' is relevant to the temperature.
In a second example (Figure~\ref{fig:example-comparison-architectures}b), only Scibert can correctly extract ``W-C nanowire'' which is not present in the SuperMat training data.
Unfortunately, we cannot check whether ``above 100K'' or ``W-C nanowire'' are also present in the dataset used in the pre-train of SciBERT by their authors~\cite{Beltagy2019SciBERT} because the data are not available.

\paragraph*{Material ML model}

To train the Material ML model we created a special dataset with an additional layer of labels (Table~\ref{tab:material-parser-entities}), which included the material information represented by entities annotated as \texttt{<material>} in the SuperMat documents.

\begin{figure}[ht]
    \centering
    \includegraphics[width=\textwidth]{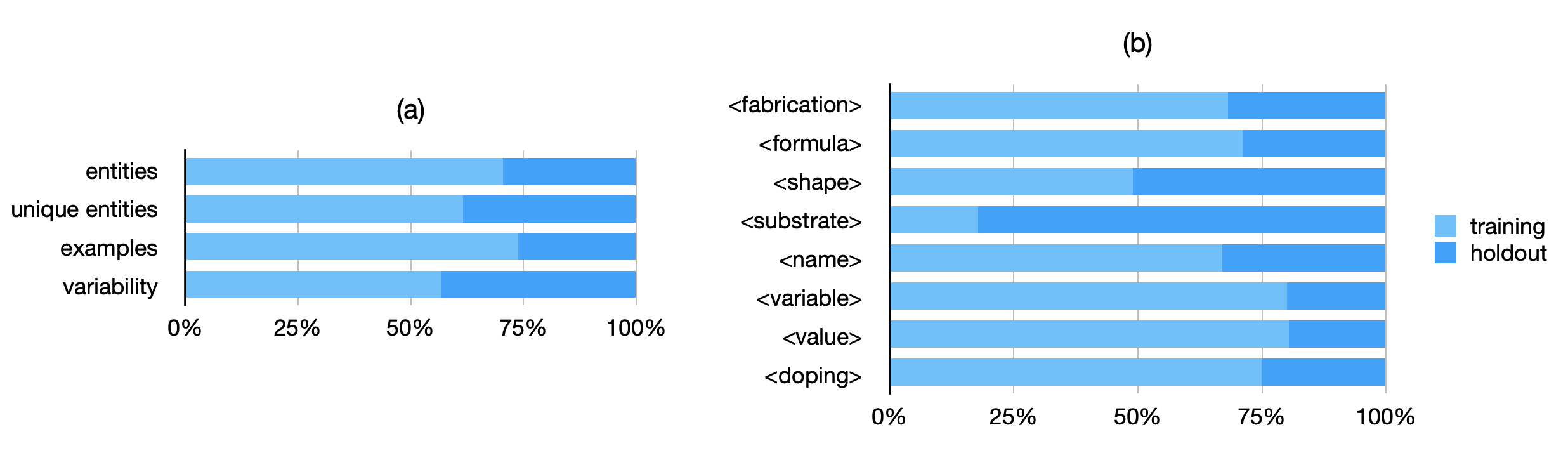}
    \caption{Holdout/training set for the Material ML model: (a) general metrics and (b) entity labels.}
    \label{fig:material-training-holdout-set-distribution}
\end{figure}

\begin{figure}[ht]
    \centering
    \includegraphics[width=0.6\textwidth]{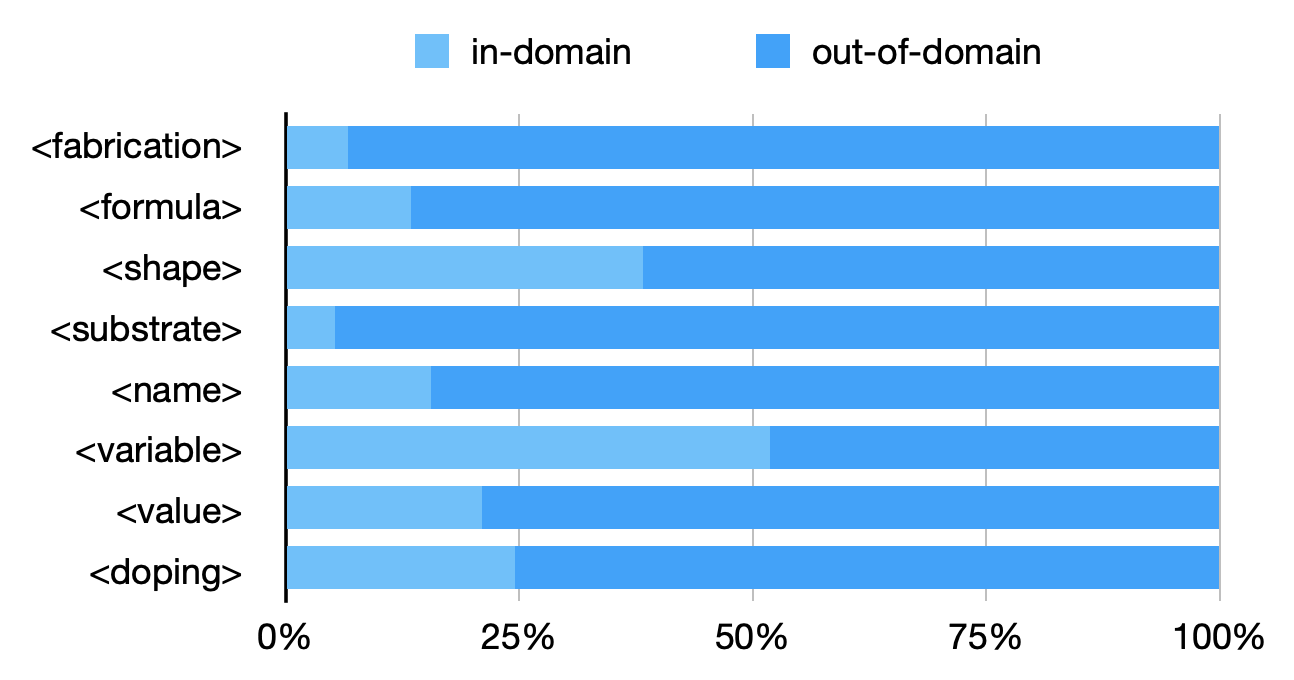}
    \caption{Holdout ``out-of-domain'' rates for the Material ML model. The entities from the holdout set that are also in the training set are the in-domain, and the entities that are not in the training set are the out-of-domain.}
    \label{fig:material-out-domain-holdout}
\end{figure}

\subparagraph*{Holdout set}
In this model we created an independent holdout set because the manual annotation work is performed on smaller chunks of text and requires less effort than annotating sentences as when we developed SuperMat.
We used material data extracted from a dataset of 500 documents (500-papers) from three publishers: \textit{American Institute of Physics} (AIP), \textit{American Physical Society} (APS) and \textit{Institute of Physics} (IOP)~\cite{foppiano2019proposal}.
The resulting holdout set has a average coverage greater than 25\% (Figure~\ref{fig:material-training-holdout-set-distribution}) and an average ``out-of-domain'' ratio of 83.93\% (Figure~\ref{fig:material-out-domain-holdout}).

\subparagraph*{Evaluation}

Scibert obtained the best results, with F1 at 84.15\% (Table~\ref{tab:evaluation-10fold-material-parser}).
The inclusion of features in the BidLSTM\_CRF architecture only improved results by less than 1\% (from 83.13 to 83.76\%).
The label \texttt{<fabrication>} did not perform well with any architecture, most likely because it is too generic (Table~\ref{tab:material-parser-entities}), and the content is too heterogeneous. Another label, \texttt{<substrate>} has only one-third of the training examples of \texttt{<fabrication>} but obtained results that were three times higher with Scibert, suggesting that \texttt{<fabrication>} should be split into separate and more homogeneous labels.

\subsection{Entity Linking}
\label{subsubsec:linking}

Entity linking (EL) links materials and their corresponding properties.

We use a rule-based algorithm, but there are other approaches such as the use of dependency parsing~\cite{yoshikawa:2017acl, Tiktinsky2020pyBARTES, swayamdipta:17, zhou-zhao-2019-head}. We did not use these because it was difficult to find a suitable dependency parser for scientific texts, and complementary methods based on complex rule sets were needed to compensate for the poor performance of the parser.

In our algorithm, pairs of entities are linked focusing on three types of link:
\begin{itemize}
    \item \textbf{material-tcValue}: The link between a material and its corresponding \tc.
    \item \textbf{tcValue-pressure}: The link between \tc and its related critical pressure.
    \item \textbf{me\_method-tcValue}: The link between \tc and its corresponding measurement method.
\end{itemize}

Entities of type \texttt{<tcValue>} are pre-processed through a classifier that establishes whether or not they temperatures related to the superconductivity. This is used to exclude other temperatures (e.g., annealing, transition, Curie) which might be incorrectly extracted by the previous step.
This rule-based classifier combines the extracted entities of \tc expressions (label \texttt{<tc>}) with a set of predefined standard terms.
If a temperature is not considered a \tc, it is excluded from the list of possible linking candidates.

Two scenarios are considered. First, if entities to be linked in the sentence are only two they are linked automatically, else further rules are applied. 
If the word ``respectively'' appears in the sentence, we apply ``order-linking''. 
For example, consider the following sentence:
\begin{displayquote}
    P-or Ba-122  and Co-doped Ba-122 have lower \tc's of about 30 K and 24 K, respectively, which makes helium free operation questionable.
\end{displayquote}
It contains the word ``respectively'', and by applying ``order-linking'', \textit{P-or Ba122} is assigned to \textit{30 K} and \textit{Co-doped Ba-122} to \textit{24 K}.


If the word ``respectively'' does not appears in the sentence, we apply ``distance-linking'' which works by defining the distance measurement \textit{d} as a value calculated as the numbers of characters between the centroid of each entity.
Entities surrounded by parenthesis are expanded to the whole parenthesis, and its centroid is updated.
As an example, in the sentence
\begin{displayquote}
    We tested two materials MgB2 (Tc = 39 K) and FeSe (Tc = 16 K).
\end{displayquote}

\texttt{39 K} is closer to \texttt{FeSe} (\textit{d}=10) than to \texttt{MgB2} (\textit{d}=11). 
In this example, however, both temperatures entities would be expanded to their containing parenthesis (e.g. ``\texttt{39 K}'' to ``\texttt{(Tc = 39 K)}''. 
In this case the centre of the entity ``\texttt{39 K}'' is shifted toward the left, from the initial value of 38 to 35 and the distance from \texttt{MgB2} is reduced from \textit{d}=11 to \textit{d}=8.
As a result, the \texttt{MgB2} entity is correctly linked to ``\texttt{39 K}''.

The distance calculation is also adjusted with the addition of ``penalties'' by doubling the calculated distance when certain keywords or punctuations (``,'', ``.'', ``;'', ``and'', ``but', ``while', ``whereas', ``which'', ``although'') appear between two entities because they represent a logical separation of predicates ~\cite{oka2021table}.
In the above example, the distance between \texttt{39 K} and \texttt{FeSe} would be doubled (\textit{d}=20) and the link would not be made.

This rule-based linking was evaluated using the linked entities from SuperMat~\cite{foppiano2021supermat} (Table~\ref{table:evaluation-linking}) and is divided considering each link type.
The F1 score for the \texttt{material-tcValue} was about 80\% with a precision of 88.40\%. 
\texttt{tcValue-pressure} F1 score was 3\% lower than  \texttt{material-tcValue} considering much less data available (support was 118 compared with 726).

\subsection{End to end evaluation}

\begin{figure}[ht]
    \centering
    \includegraphics[width=\textwidth]{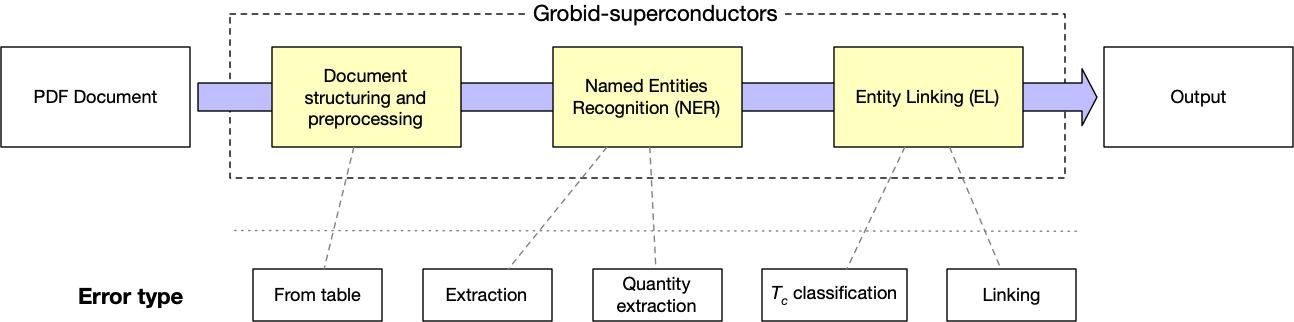}
    \caption{\textit{Error types} in the context of the data flow. }
    \label{fig:error-types}
\end{figure}

End-to-end evaluation (E2EE) measures the capacity of the system from the PDF documents until the final linked results.
We limited the scope of the E2EE to the triplet `material-\tc-pressure' which, at the moment, is the backbone upon which the database is built.
We performed the E2EE on the ``500-papers'' dataset where we manually examined the resulting database as follows: 1) we marked invalid records and 2) we identified the cause of failure from a predefined set of five \textit{error types} (Figure~\ref{fig:error-types}):
\begin{itemize}
    \item \textbf{From table}: the extracted text is wrongly extracted from a table. Although table content is ignored, the error rate from the Grobid library is still relevant due to the lack of training data.
    \item \textbf{Extraction}: entities are not recognised, wrongly recognised, or partially recognised.
    \item \textbf{Quantity extraction}: quantity entities (pressure, temperature) are not correctly extracted. We measured this error separately to identify the failure that could be shared with the Quantity ML model.
    \item \textbf{\tc classification}: the temperature is wrongly classified as superconducting \tc.
    \item \textbf{Linking}: given the initial steps were performed correctly, the resulting entities are not linked correctly.
\end{itemize}

The E2EE scores are summarised in Table~\ref{table:end2end-evaluation-summary}.
Recall is omitted because it is less relevant and difficult to calculate manually.
The precision score (micro average) was 72.60\% for all the subsections, although the error rates of figure captions (59.28\%) and unknown subsections (57.14\%) were clearly lower than those of the other subsections ($>$ 70\%).
The `unknown` subsections indicate that the extracted text's structure was not well identified by Grobid but it was nevertheless aggregated.
The overall score increases to 73\% when excluding unknown subsections, 75.24\% when excluding figure captions, and 79.14\%  when excluding both.
Excluding these two subsections will not impact the amount of text, because both account for less than 20\% of the total number of subsections.

\begin{figure}[ht]
    \centering
    \includegraphics[width=\linewidth]{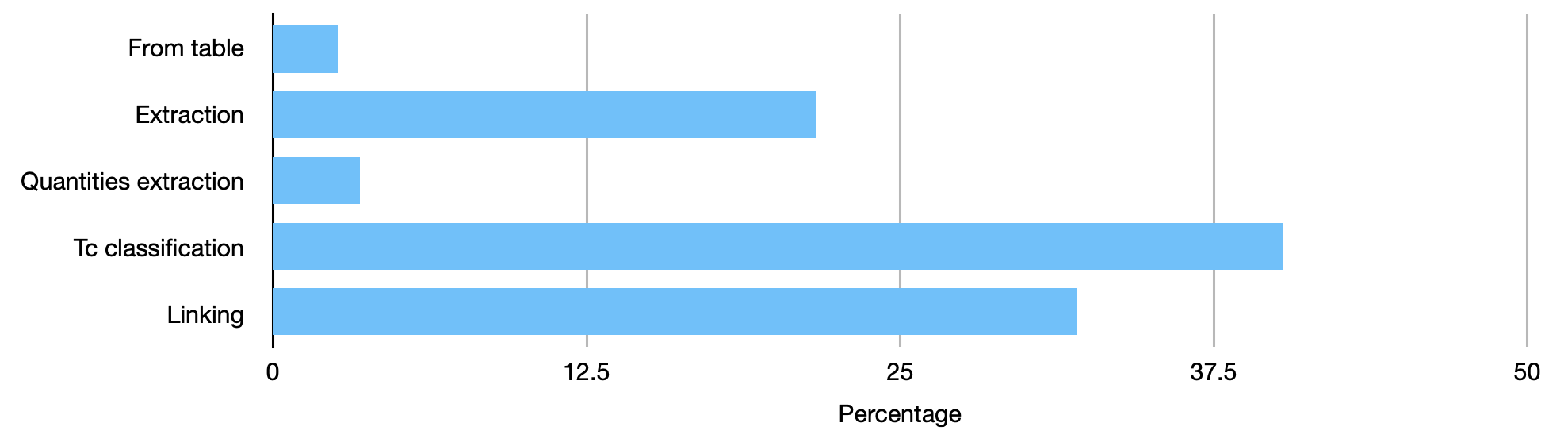}
    \caption{Error type distribution in the E2EE of the \textit{500-papers} dataset.}
    \label{fig:error-types-distribution}
\end{figure}

The error types are summarised in Figure~\ref{fig:error-types-distribution}. The most common failures  originate from \tc~classification (40\%), Linking (32\%), and Extraction (20\%).
The most common \tc classification failures are as incorrect recognition of 1) relative values of \tc (e.g., 1 K higher than material X); 2) values indicating the transition temperature width ($\Delta T_{c}$); 3) temperature values that are not \tc, for example, material synthesis temperatures ($T$), other critical transition temperatures that are not superconducting (e.g., $T_{Curie}$); and 4) values of temperature at which there is no superconductivity (e.g., ``at 70 K there is no superconductivity'').
``Linking errors'' mainly occur when the text compare relative values of \tc~using materials as the basis for comparison (e.g., ``The Tc = 38 K is similar to the one of MgB$_{2}$'').
Finally, ``Extraction'' issues mainly originate from: 1) implicit mention of the main material when experimented using different ``substrates'' combination, and 2) mismatches between \texttt{<material>} and \texttt{<class>} which, by definition, overlap.


\section{Supercon\textsuperscript{2}}

We created SuperCon\textsuperscript{2} by processing 37770 research papers belonging to the category \textit{cond-mat.supr-cond} in ArXiv.
Currently SuperCon\textsuperscript{2} contains 40324 records including 2052 triplets with applied pressure (\textit{material-\tc-pressure}), and 3602 records with explicit measurement method (\textit{material-\tc-measurement method}).
The schema of SuperCon\textsuperscript{2} is summarised with examples in Table~\ref{tab:supercon2-schema}.

\begin{figure}[ht]
    \centering
    \includegraphics[width=0.8\textwidth]{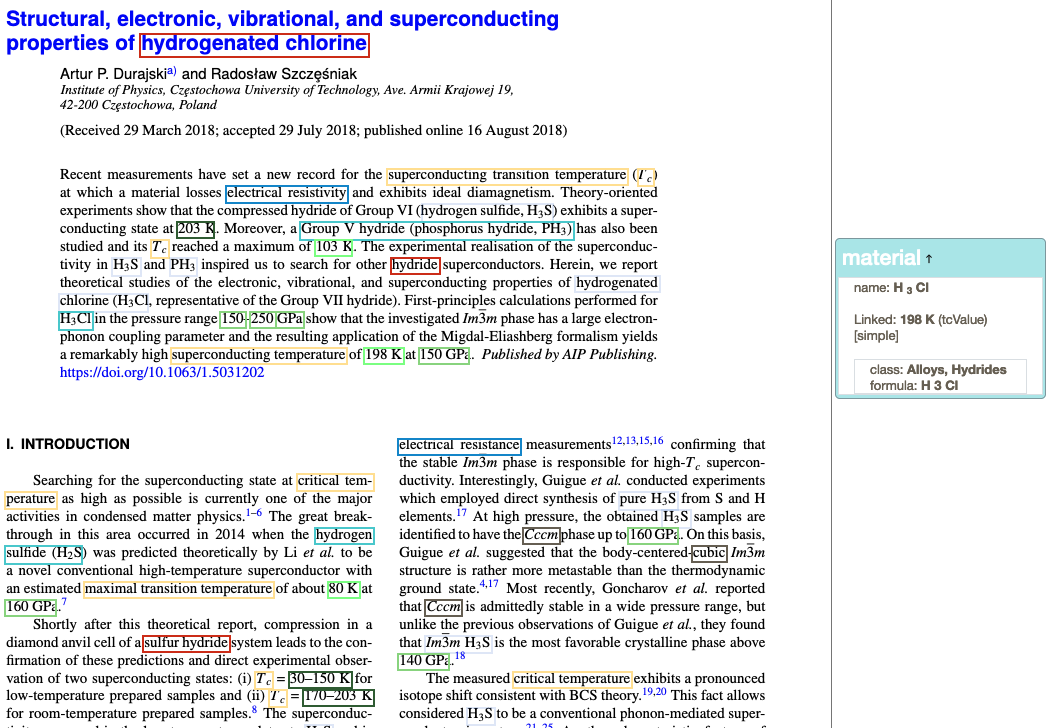}
    \caption{\label{fig:pdf-annotations} Example of a superconductors research PDF document~\cite{sample_superconductors_article} enriched with extracted annotations. Materials information (class, formula) and properties (\tc) are summarised in the information box when the users click on the highlighted annotated entity in the text.}
\end{figure}

The data is processed and ingested through the asynchronous Map-Reduce approach~\cite{10.1145/1327452.1327492}.
The ``extraction task'' (Map) processes the PDF documents by accessing Grobid-superconductors via REST API and stores their processed representation together with the original PDF document.
Furthermore, the ``aggregation task'' (Reduce) reduces the document information into a synthesised tabular format.
We store the processed document representation in JSON format. 
The processed documents are kept separately and used for displaying the enhanced PDF document (Figure~\ref{fig:pdf-annotations}).
The pipeline uses a persistence layer for storage and reporting (logger).



We built a visualisation interface to exploit the extracted information.
Users can search in the synthesised tabular data, access the PDF document enriched with the extracted information (Figure~\ref{fig:pdf-annotations}), and export locally in CSV, TSV and Microsoft Excel formats.

\section{Conclusion}
\label{sec:conclusion}
In this work, we present our solution for automatically building a database of materials and properties from scientific literature.
Our contribution is composed of: 1) Grobid-superconductors, a specialised open-source system that processes PDF documents combining ML and rule-based methods to extract and link relevant information in superconductors research; 2) a pipeline allowing large-scale document processing; and 3) a visualisation interface for rapid data exploration, which includes PDF document information enrichment.

We made SuperCon\textsuperscript{2}, a database with 40324 records of superconductors materials and properties, including the applied pressure and the \tc~measurement method.
SuperCon\textsuperscript{2} is available in text format at \url{https://github.com/lfoppiano/supercon}.

In the future, we plan to improve our tools by 1) extracting more properties, such as crystal structure type, space groups type, and lattice structure; 2) training supervised models for the ``Linking step''; and 3) extending the interface to support data correction toward efficient curation.
We confirmed the good generalisation ability of the Scibert architecture for the entity extraction task.
Although we hope to obtain better results using materials science pre-trained BERT, such as MatSciBERT~\cite{gupta_matscibert_2022}, the gain might be just minimal for relatively larger models~\cite{hong2022ScholarBERT}.

\section*{Acknowledgement}
\label{sec:acknowledgement}
Our warmest thanks to Sae Dieb, who contributed with fruitful discussion on this project, and to Patrice Lopez, the author of Grobid~\cite{GROBID}, DeLFT~\citep{DeLFT}, and other several interesting TDM projects for his continuous support and help with ideas, suggestions, and discussions.


\section*{Data and code availability}
Grobid-superconductors is available on Github at \url{https://github.com/lfoppiano/grobid-superconductors} and the code is released under license Apache 2.0. 
SuperCon\textsuperscript{2} is available in text format at \url{https://github.com/lfoppiano/supercon}.

\section*{Competing interests}
The authors declare no competing interests.

\bibliography{bibliography}
\bibliographystyle{tfnlm}

\section*{Figures \& Tables}

\begin{table}[ht]
    \centering
    \tbl{Results from cross-validation for sentence-based and paragraphs-based text.  }
    {
        \begin{tabular}{lrrr}
            \toprule
            \textbf{Label}             & \textbf{Precision} & \textbf{Recall} & \textbf{F1} \\
            \midrule
            Paragraph-based micro avg. & 44.44              & 27.21           & 33.76       \\
            Sentence-based micro avg.  & 48.41              & 50.00           & 51.70       \\
            \bottomrule
        \end{tabular}
    }
    \label{tab:comparison-evaluation-sentences-paragraphs}
\end{table}

\begin{table}[ht]
    \tbl{Entities extracted by the superconductors parser.}
    {\begin{tabular}{m{19em} m{30em}}
        \toprule
        \textbf{Entity} (\textbf{tag})              & \textbf{Description} \\
        \midrule
        Material (\texttt{<material>})              & Materials and samples names, formulas (including stochiometric formulas), substitution variables of values and elements, shape, doping, and substrate               \\
        Class (\texttt{<class>})                    & Groups of materials having similar characteristics or common strategic compounds that define their nature                                                      \\
        \tc value (\texttt{<tcValue>})      & The value of the superconductor critical temperature                                                                                                          \\
        \tc expressions (\texttt{<tc>})     & Expressions in the text that provide information about the phenomenon of superconductivity related to a value, interval or variation of the \tc \\
        Measurement method (\texttt{<me\_method>}) & Technique used to measure or calculate the presence of superconductivity                                                                                     \\
        Applied pressure (\texttt{<pressure>})      & Applied pressure when superconductivity is recorded                                                                                                            \\
        \bottomrule
    \end{tabular}}
    \label{tab:superconductors-parser-entities}
\end{table}

\begin{table}[ht]
    \tbl{Entities extracted by the material parser. }
    {
        \begin{tabular}{m{16em} m{30em}}
            \toprule
            \textbf{Entity} (\textbf{tag})               & \textbf{Description}                                                                                                              \\
            \midrule
            Name (\texttt{<name>})                       & The canonical name of a material (e.g., hydrogen, PCCO, carbon)                                                                    \\
            Formula (\texttt{<formula>})                 & Chemical formula of the material (e.g., \texttt{Pr1.869Ce0.131CuO 4-}, \texttt{MgB2}, \texttt{La 2-x Sr x CuO 4})                  \\
            Doping (\texttt{<doping>})                   & Doping ratio and doping materials that are adjoined to the material name (e.g., \texttt{Zn-doped}, \texttt{2\% Zn-doped})          \\
            Shape (\texttt{<shape>})                     & shape of the material (e.g. single crystal, polycrystalline, thin film, powder, film)                                             \\
            Substitution variables (\texttt{<variable>}) & Variables that can be substituted in the formula.                                                                                 \\
            Substitution values (\texttt{<value>})       & Values expressed in the doping.                                                                                                   \\
            Substrate (\texttt{<substrate>})             & Substrates as defined in the material name                                                                                        \\
            Fabrication (\texttt{<fabrication>})         & Additional information that does not belong to any of the previous tags  (e.g., intercalated, electron-doped) \\
            \bottomrule
        \end{tabular}
    }
    \label{tab:material-parser-entities}
\end{table}

\begin{table}[ht]
    \tbl{Evaluation scores (\%) for the Superconductor ML model in the four architectures. For the DL architecture the results are averaged over 5 runs. Support (Supp) indicates the number of labels in the training data. Values in bold indicate the highest score. P: precision, R: recall.}
    {
    \scalebox{0.8}{
        \begin{tabular}{l ccc ccc ccc ccc r}
            \toprule
            \textbf{Label}        & \multicolumn{3}{c}{\textbf{CRF}} & \multicolumn{3}{c}{\textbf{BidLSTM\_CRF}} & \multicolumn{3}{c}{\makecell{\textbf{BidLSTM\_CRF}                                                                                                                                                 \\\textbf{\_FEATURES}}} & \multicolumn{3}{c}{\textbf{Scibert}} & \textbf{Supp} \\
            \cmidrule(lr){2-4}\cmidrule(lr){5-7}\cmidrule(lr){8-10}\cmidrule(lr){11-13}\cmidrule(lr){14-14}
                                  & \textbf{P}                       & \textbf{R}                                & \textbf{F1}                                        & \textbf{P} & \textbf{R} & \textbf{F1}    & \textbf{P}     & \textbf{R} & \textbf{F1}    & \textbf{P} & \textbf{R}     & \textbf{F1}    &      \\
            \midrule
            \texttt{<class>}      & 79.74                            & 66.79                                     & 72.69                                              & 79.01      & 72.62      & \textbf{75.66} & 77.84          & 72.40      & 74.97          & 72.95      & 75.28          & 74.09          & 1646 \\
            \texttt{<material>}   & 79                               & 72.15                                     & 75.42                                              & 79.25      & 76.94      & 78.06          & 81.07          & 75.10      & 77.94          & 80.15      & 81.42          & \textbf{80.77} & 6943 \\
            \texttt{<me\_method>} & 60.25                            & 68.73                                     & 64.21                                              & 56.41      & 79.49      & 65.92          & 55.86          & 80.45      & 65.90          & 56.26      & 81.52          & \textbf{66.56} & 1883 \\
            \texttt{<pressure>}   & 46.15                            & 29.27                                     & 35.82                                              & 49.45      & 58.05      & 52.53          & 50.25          & 60.49      & \textbf{54.36} & 41.72      & 52.68          & 46.51          & 274  \\
            \texttt{<tc>}         & 84.36                            & 83.57                                     & \textbf{83.96}                                     & 78.61      & 82.54      & 80.48          & 79.19          & 82.07      & 80.60          & 74.46      & 82.66          & 78.35          & 3741 \\
            \texttt{<tcValue>}    & 69.8                             & 66.24                                     & 67.97                                              & 70.36      & 75.16      & 72.67          & 68.95          & 76.56      & 72.52          & 70.90      & 79.74          & \textbf{75.06} & 1099 \\
            \midrule
            All (micro avg)       & 76.88                            & 72.77                                     & 74.77                                              & 74.59      & 77.67      & 76.09          & \textbf{75.17} & 76.79      & 75.96          & 73.69      & \textbf{80.69} & \textbf{77.03}        \\
            \bottomrule
        \end{tabular}
    }
    }
    \label{tab:evaluation-superconductors-ML-model} 
\end{table}

\begin{table}[ht]
    \centering\small
    \tbl{Evaluation scores (\%) of the Material ML model with holdout set. Support (Supp) indicates the number of labels in the training data. Values in bold indicate the highest score. P: precision, R: recall.}
    {
        \scalebox{0.8}{
        \begin{tabular}{l ccc ccc ccc ccc r}
            \toprule
            \textbf{Label}         & \multicolumn{3}{c}{\textbf{CRF}} & \multicolumn{3}{c}{\textbf{BidLSTM\_CRF}} & \multicolumn{3}{c}{\makecell{\textbf{BidLSTM\_CRF}                                                                                                                                                 \\\textbf{\_FEATURES}}} & \multicolumn{3}{c}{\textbf{SciBERT}} & \textbf{Supp}  \\
            \cmidrule(lr){2-4}\cmidrule(lr){5-7}\cmidrule(lr){8-10}\cmidrule(lr){11-13}\cmidrule(lr){14-14}
                                   & \textbf{P}                       & \textbf{R}                                & \textbf{F1}                                        & \textbf{P} & \textbf{R} & \textbf{F1}    & \textbf{P}     & \textbf{R} & \textbf{F1}    & \textbf{P} & \textbf{R}     & \textbf{F1}    &      \\
            \midrule
            \texttt{<doping>}      & 60.41                            & 55.85                                     & 58.04                                              & 67.98      & 62.42      & 64.95          & 69.00          & 62.34      & \textbf{65.43} & 63.58      & 62.79          & 63.16          & 792  \\
            \texttt{<fabrication>} & 40.00                            & 4.55                                      & 8.16                                               & 23.61      & 5.91       & 9.24           & 37.33          & 9.09       & 14.48          & 22.51      & 13.18          & \textbf{16.52} & 94   \\
            \texttt{<formula>}     & 80.81                            & 82.29                                     & 81.54                                              & 82.59      & 84.14      & 83.35          & 83.83          & 85.14      & 84.47          & 84.53      & 86.56          & \textbf{85.53} & 6301 \\
            \texttt{<name>}        & 72.2                             & 63.75                                     & 67.71                                              & 76.29      & 78.76      & 77.43          & 74.51          & 80.38      & 77.33          & 77.18      & 81.86          & \textbf{79.44} & 1930 \\
            \texttt{<shape>}       & 90.89                            & 92.51                                     & 91.69                                              & 90.93      & 95.79      & \textbf{93.29} & 90.33          & 95.74      & 92.96          & 89.67      & 97.20          & 93.28          & 809  \\
            \texttt{<substrate>}   & 37.04                            & 6.76                                      & 11.43                                              & 54.31      & 32.43      & 40.44          & 60.08          & 33.38      & 42.82          & 56.32      & 41.22          & \textbf{47.59} & 32   \\
            \texttt{<value>}       & 80.21                            & 83.15                                     & 81.65                                              & 84.81      & 89.33      & 86.99          & 85.16          & 90.15      & \textbf{87.58} & 83.14      & 85.92          & 84.50          & 1895 \\
            \texttt{<variable>}    & 96.85                            & 95.98                                     & 96.41                                              & 95.19      & 97.77      & 96.46          & 96.32          & 97.90      & \textbf{97.10} & 96.22      & 96.52          & 96.37          & 1795 \\
            \midrule
            All (micro avg)        & 81.15                            & 78.09                                     & 79.59                                              & 82.76      & 83.50      & 83.13          & \textbf{83.20} & 84.33      & 83.76          & 83.11      & \textbf{85.23} & \textbf{84.15} &      \\
            \bottomrule
        \end{tabular}
    }
    }
    \label{tab:evaluation-10fold-material-parser}
\end{table}

\begin{table}[ht]
    \centering
    \tbl{Evaluation scores for the Linking. Support (Supp) indicates the number of labels in the training data. Values in bold indicate the highest score. P: precision, R: recall.}{
        \begin{tabular}{lcccc}
            \toprule
            \textbf{Relationship type}          & \textbf{P} & \textbf{R} & \textbf{F1-} & Supp \\
            \midrule
            \textbf{material-tcValue}   & 88.40              & 74.52           & 80.87             & 726     \\
            \textbf{tcValue-pressure}   & 85.71              & 71.52           & 77.98             & 118     \\
            \textbf{me\_method-tcValue} & 62.28              & 65.74           & 63.96             & 151     \\
            \bottomrule
        \end{tabular}
    }
    \label{table:evaluation-linking}
\end{table}

\begin{table}[ht]
    \centering
    \tbl{Summary of the E2EE evaluation scores. Support indicates the number of labels in the training data.}{
        \begin{tabular}{l c c}
            \toprule
            \textbf{Subsection}                                      & \textbf{Precision} & \textbf{Support} \\
            \midrule
            Title                                                    & 100                & 2                \\
            Abstract                                                 & 80.32              & 61               \\
            Paragraph                                                & 75.2               & 623              \\
            Figure captions                                          & 59.28              & 140              \\
            Unknown                                                  & 57.14              & 21               \\
            \midrule
            \textbf{Micro avg.}                                      & 72.60              & 847              \\
            \textbf{Micro avg.} (excl. figures)                      & 75.24              & 707              \\
            \textbf{Micro avg.} (excl. unknown sections)             & 73.00              & 603              \\
            \textbf{Micro avg.} (excl. figures and unknown sections) & 79.14              & 657              \\
            \bottomrule
        \end{tabular}
    }
    \label{table:end2end-evaluation-summary}
\end{table}

\afterpage {
    \clearpage 

    \begin{table}[ht]
        \centering
        \tbl{Summary and description of the SuperCon\textsuperscript{2} schema. ``Internal information'' is technical information not accessible to the users.}{
            \begin{tblr}{Q[l,m]Q[r,m]Q[r,m]}
                \hline[1pt]
                \textbf{Field name} & \textbf{Description}                             & \textbf{Examples}             \\
                \hline
                \multicolumn{3}{c}{\emph{Material information}}                                                        \\
                \hline[dashed]
                {Raw                                                                                                   \\ material} & The material or sample as it appears in the text &\\
                \hline[dotted]
                Name                & Canonical name of a material                     & {PCCO, PCO, Metal diboride,   \\ hydrogen, carbon} \\
                \hline[dotted]
                Formula             & {Material expressed as chemical formula. This                                    \\ includes also formulas with stochiometric variables} & {$Pr_{1.869}Ce_{0.131}CuO_4-\delta$,\\ $MgB_2$, $La_{2-x} Sr_x CuO_4$} \\
                \hline[dotted]
                Doping              & {Doping ratio and doping materials                                               \\ that might be adjoined to the material} & {Overdoped, underdopded,\\ optimally doped,\\ bulk, pure, 1\% Zn, Zn\\ (from Zn-doped XYZ)}\\
                \hline[dotted]
                Shape               & The shape of the material or the sample          & {Single crystal, polycrystal, \\ wire, powder, film} \\
                \hline[dotted]
                Variables           & Variables that can be substituted in the formula & x = 0, RE=Ln,St               \\
                \hline[dotted]
                Class               & {Material classification according                                               \\ to the domain-experts taxonomy} & cuprates, oxides, and alloys\\
                \hline[dotted]
                Fabrication         & {All the information that does not                                               \\ belong to any of the previous tags} &  {Intercalated,\\ synthesized by MBE method,\\ electron-doped, hole-doped} \\
                \hline[dotted]
                Substrate           & Substrate material described in the raw material & {PCCO films onto              \\ $Pr_2 CuO_4 (PCO)/SrTiO_3$ }\\
                \hline[dashed]
                \multicolumn{3}{c}{\emph{Properties}}                                                                  \\
                \hline[dashed]
                {Critical                                                                                              \\ Temperature}  & Superconducting critical temperature &\\
                \hline[dotted]
                {Applied                                                                                               \\ Pressure}  & {Pressure applied when measuring \\ the superconducting critical temperature} &\\
                \hline[dotted]
                {Measurement                                                                                           \\ Method}  & {Method for measurement of the\\ superconducting critical temperature} & {Magnetic susceptibility,\\ specific heat, calculation,\\ prediction, resistivity}\\
                \hline[dashed]
                \multicolumn{3}{c}{\emph{Document bibliographic information}}                                          \\
                \hline[dashed]
                Section             & The main body section of the paper               & Header, body, annex           \\
                \hline[dotted]
                Subsection          & The secondary segmentation area of the paper     & {Paragraph, table caption,    \\ figure caption, title, abstract} \\
                \hline[dotted]
                {Authors,                                                                                              \\ Title, DOI,\\ Publisher,\\ Journal, Year} & \multicolumn{2}{c}{Bibliographic information of the document}\\
                \hline[dashed]
                \multicolumn{3}{c}{\emph{Internal information}}                                                        \\
                \hline[dashed]
                {Hash,                                                                                                 \\ Timestamp} & \multicolumn{2}{c}{Hash calculated on the binary content of the original PDF\\ document and the timestamp when the document was processed.}\\
                \hline[1pt]
            \end{tblr}
        }
        \label{tab:supercon2-schema}
    \end{table}
    \clearpage
}

\clearpage

\appendix

\section{Additional information}

\begin{table}[ht]
    \centering
    \tbl{Holdout/Training set distribution (\%) between training and holdout sets for the Superconductor ML model. Positive examples indicate the number of sentences with at least one entity, and negative examples the number of sentences with no entities.}
    {
        \begin{tabular}{lccc}
            \toprule
                                          & \textbf{training} & \textbf{holdout} & \textbf{\% holdout/training} \\
            \midrule
            \textbf{documents}         & 132               & 32               & 24.24\%                   \\
            \textbf{examples}          & 16902             & 3905             & 23.10\%                   \\
            \textbf{entities}          & 15586             & 3112             & 19.97\%                   \\
            \textbf{unique entities}   & 6699              & 1372             & 20.48\%                   \\
            \textbf{positive examples} & 8380              & 1776             & 21.19\%                   \\
            \textbf{negative examples} & 8522              & 2129             & 24.98\%                   \\
            \bottomrule
        \end{tabular}
    }
    \label{tab:training-holdout-set-distribution-annex}
\end{table}

\begin{table}[ht]
    \centering
    \tbl{Holdout/Training set distribution (\%) between training and holdout sets on different labels for the Superconductors ML model.}{
    \begin{tabular}{lccc}
        \toprule
        label                 & \textbf{training} & \textbf{holdout} & \textbf{\% holdout/training } \\
        \midrule
        \texttt{<tc>}         & 3741              & 639              & 17.08\%                    \\
        \texttt{<class>}      & 1646              & 271              & 16.46\%                    \\
        \texttt{<material>}   & 6943              & 1649             & 23.75\%                    \\
        \texttt{<me\_method>} & 1883              & 355              & 18.85\%                    \\
        \texttt{<tcValue>}    & 1099              & 157              & 14.29\%                    \\
        \texttt{<pressure>}   & 274               & 41               & 14.96\%                    \\
        \bottomrule
    \end{tabular}
    }
    \label{tab:training-holdout-labels-distribution-annex}
\end{table}

\begin{table}[ht]
    \centering
    \tbl{Holdout/Training set distribution (\%) training and holdout sets for the Material ML model.}{
        \begin{tabular}{lccc}
            \toprule
                                        & \textbf{training} & \textbf{holdout} & \textbf{\% holdout/training} \\
            \midrule
            \textbf{examples}        & 13648             & 5728             & 41.97\%                   \\
            \textbf{entities}        & 4512              & 2817             & 62.43\%                   \\
            \textbf{unique entities} & 9268              & 3292             & 35.52\%                   \\
            \bottomrule
        \end{tabular}
    }
    \label{tab:training-holdout-set-material-distribution-annex}
\end{table}

\begin{table}[ht]
    \centering
    \tbl{Holdout/Training set distribution (\%) training and holdout sets on different labels for the Material ML model.}{
    \begin{tabular}{lccc}
        \toprule
        label                  & \textbf{training} & \textbf{holdout} & \textbf{\% holdout/training } \\
        \midrule
        \texttt{<fabrication>} & 94                & 44               & 46.81\%                    \\
        \texttt{<formula>}     & 6301              & 2569             & 40.77\%                    \\
        \texttt{<shape>}       & 809               & 841              & 103.96\%                   \\
        \texttt{<substrate>}   & 32                & 148              & 462.50\%                   \\
        \texttt{<name>}        & 1930              & 949              & 49.17\%                    \\
        \texttt{<variable>}    & 1795              & 449              & 25.01\%                    \\
        \texttt{<value>}       & 1895              & 463              & 24.43\%                    \\
        \texttt{<doping>}      & 792               & 265              & 33.46\%                    \\
        \bottomrule
    \end{tabular}
    }
    \label{tab:training-holdout-labels-material-distribution-annex}
\end{table}

\begin{table}[ht]
    \centering
    \tbl{Summary of the features used in the \textit{superconductors} and \textit{material} ML models. \textit{All} under Architecture indicate only BidLSTM\_CRF\_FEATURES and CRF.}{
        \begin{tabular}{l m{30em} c c}
            \toprule
            \textbf{\#}   & \textbf{Feature}                                                                                                                                                                                                                                         & \textbf{Model}  & \textbf{Architecture} \\
            \midrule
            \textbf{1}    & current token                                                                                                                                                                                                                                            & all             & all                   \\
            \textbf{2}    & current token lower cased                                                                                                                                                                                                                                & all             & all                   \\
            \textbf{3-6}  & (four features) current token, prefix characters 1 to 4                                                                                                                                                                                                  & all             & CRF                   \\
            \textbf{7-10} & (four features) current token, suffix characters 1 to 4                                                                                                                                                                                                  & all             & CRF                   \\
            \textbf{11}   & information about capitalisation: first character (INITCAP), all characters (ALLCAPS), none (NOCAPS)                                                                                                                                                     & all             & all                   \\
            \textbf{12}   & digits content: all (ALLDIGIT), some digits (CONTAINDIGIT), no digits (NODIGIT)                                                                                                                                                                          & all             & all                   \\
            \textbf{13}   & (boolean) the token is composed by a single character                                                                                                                                                                                                    & all             & all                   \\
            \textbf{14}   & punctuaction information and normalisation to placeholders: no punctuation (NOPUNCT), open or end brackets (OPENBRACKET, ENDBRACKET), various punctuation (DOT, COMMA, HYPHEN, QUOTE), open or close quotes (OPENQUOTE, ENDQUOTE), anything else (PUNCT) & all             & all                   \\
            \textbf{15}   & Shadow the numbers                                                                                                                                                                                                                                       & all             & CRF                   \\
            \textbf{16}   & Shadow any characters: ``x'' for lowercase, ``X'' for uppercase, ``d'' for digits                                                                                                                                                                        & all             & CRF                   \\
            \textbf{17}   & As the previous but compressed                                                                                                                                                                                                                           & all             & CRF                   \\
            \textbf{18}   & Font name                                                                                                                                                                                                                                                & superconductors & all                   \\
            \textbf{19}   & Font size                                                                                                                                                                                                                                                & superconductors & all                   \\
            \textbf{20}   & Font style: standard (BASELINE), superscript (SUPERSCRIPT) or subscript (SUBSCRIPT)                                                                                                                                                                      & superconductors & all                   \\
            \textbf{21}   & (boolean) if the token style is bold                                                                                                                                                                                                                     & superconductors & all                   \\
            \textbf{22}   & (boolean) if the token style is italic                                                                                                                                                                                                                   & superconductors & all                   \\
            \textbf{23}   & (boolean) the token is identified as a chemical compound by ChemDataExtractor\cite{chemdataextractor}                                                                                                                                                    & superconductors & all                   \\
            \bottomrule
        \end{tabular}
    }
    \label{tab:ML-model-features}
\end{table}


\end{document}